\documentclass[conference]{IEEEtran}
\IEEEoverridecommandlockouts
\usepackage{cite}
\usepackage{amsmath,amssymb,amsfonts}
\usepackage{algorithmic}
\usepackage{graphicx}
\usepackage{textcomp}
\usepackage{xcolor}
\usepackage{booktabs}
\usepackage{subcaption}  
\usepackage{multirow}
\usepackage{newtxtext,newtxmath}
\usepackage{pdfx}
\def\BibTeX{{\rm B\kern-.05em{\sc i\kern-.025em b}\kern-.08em
    T\kern-.1667em\lower.7ex\hbox{E}\kern-.125emX}}
\begin{document}

\title{
Enhancing Multi-modal Models with Heterogeneous MoE Adapters for Fine-tuning}

\author{
    \IEEEauthorblockN{Sashuai Zhou}
    \IEEEauthorblockA{
        College of Computer Science \\
        Zhejiang University, China \\
        chouss911@gmail.com	
    }
    \and
    \IEEEauthorblockN{Hai Huang}
    \IEEEauthorblockA{
        College of Software \\
        Zhejiang University, China \\
        haihuangcode@outlook.com	
    }
    \and
    \IEEEauthorblockN{Yan Xia\textsuperscript{\dag}}
    \IEEEauthorblockA{
        College of Computer Science \\
        Zhejiang University, China \\
        xiayan.zju@gmail.com	
    }
    \thanks{\textsuperscript{\dag} Corresponding author}
}

\maketitle

\begin{abstract}

Multi-modal models excel in cross-modal tasks but are computationally expensive due to their billions of parameters. Parameter-efficient fine-tuning (PEFT) offers a solution by adding small trainable components while freezing pre-trained parameters. However, existing  methods primarily focus on uni-modal processing, overlooking the critical modal fusion needed for multi-modal tasks. To fill this gap, we propose heterogeneous mixture of experts adapters that extend the traditional PEFT framework to support multi-modal expert combinations and improve information interaction. Additionally, our approach modifies the affine linear expert design to enable efficient modal fusion in a low-rank space, achieving competitive performance with only 5-8\% of the parameters fine-tuned. Experiments across eight downstream tasks, including visual-audio and text-visual, demonstrate the superior performance of the approach.

\end{abstract}


\section{Introduction}

The world is inherently multi-modal, with humans perceiving information through diverse sensory modalities such as language, images, and sounds. Recent advancements in large language models (LLMs) \cite{chowdhery2023palm, achiam2023gpt}have enabled them to process not only text but also vision, video, and audio, significantly enhancing their performance in applications like search engines and intelligent assistants. However, fine-tuning multi-modal LLMs remains computationally expensive\cite{he2022unified}, posing challenges for broader accessibility and scalability.

Parameter-Efficient Fine-Tuning (PEFT) \cite{ houlsby2019parameterefficient,zhang2023lorafa,Luo2023TowardsEV}techniques reduce fine-tuning costs by adding small trainable components while freezing the original model parameters. While most PEFT methods focus on single-modality tasks and lack effective mechanisms for multi-modal fusion, limiting their performance in complex interactions. A further advance in this area is the introduction of Mixture of Experts (MoE)-based adapters\cite{6797059,Eigen2013LearningFR,shazeer2017outrageously,mustafa2022multimodal}, which incorporate multiple adapters within transformer layers and use a router to select the optimal expert combination for each task. This approach enhances model capacity while maintaining inference efficiency. However, existing MoE adapters typically rely on simple two-layer structures and process each modality separately, limiting their effectiveness in complex multi-modal tasks like visual-audio fusion\cite{xuan2020cross,tian2020unified,li2022learning,zhou2022audio}. Specifically, these methods suffer from two main issues: 1) They treat modalities independently, neglecting the essential cross-modal interactions needed for downstream tasks, and 2) Freezing the original model parameters hampers effective multi-modal interaction within the trainable layers, restricting the model’s full potential.


\begin{figure*}[t]
    \centering
    \includegraphics[width=\textwidth]{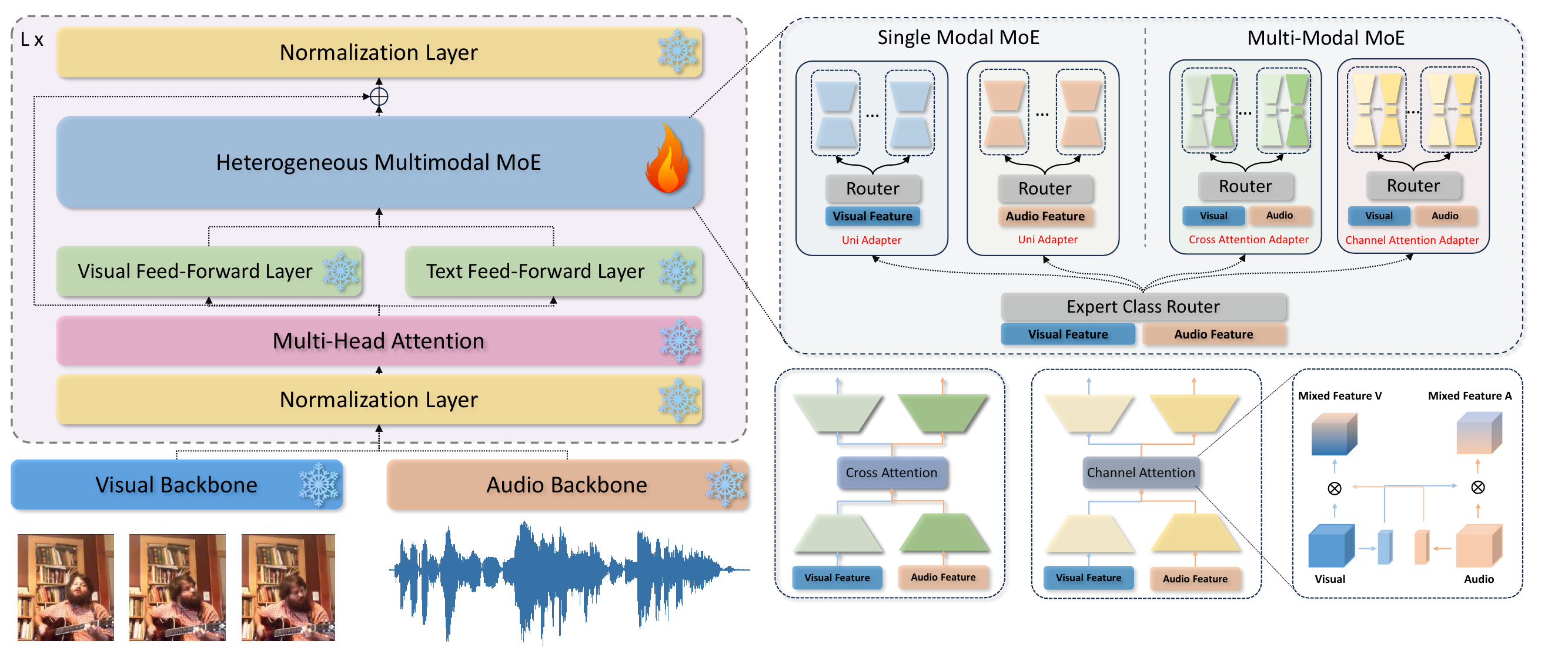}
    \caption{The overall architecture of our proposed method. The left half shows the heterogeneous multi-modal MoE inserted into
the pre-trained model as an additional trainable layer. The upper part of the right half shows the overall routing structure. The
bottom half on the right shows the internal structure of the heterogeneous experts.}
    \label{fig:aaai}
\end{figure*}

To enhance modal interactions during multi-modal model fine-tuning, we introduce the Heterogeneous Multi-Modal Mixture of Experts Adapter (HMMoE). This approach allows each expert to process inputs from multiple modalities, enabling effective cross-modal fusion. Furthermore, we replace the traditional single-expert structure with a heterogeneous architecture that combines conventional adapters with specialized multi-modal interaction experts, such as cross-attention experts for capturing inter-modal dependencies and channel-attention experts for targeted feature extraction. Experts are grouped by type, with each group comprising multiple identical adapters. 

We integrate the proposed HMMoE modules into existing multi-modal models and conduct extensive experiments on visual-audio and text-vision tasks. Experimental results demonstrate that our method achieves performance comparable to full fine-tuning while utilizing only 5-8\% of the parameters. Additionally, it significantly surpasses existing Parameter-Efficient Fine-Tuning methods, providing an effective solution for fine-tuning multi-modal models with minimal parameter overhead. Our main contributions are as follows: 
\begin{itemize} \item We propose the HMMoE module to enhance cross-modal understanding through interactions in a low-dimensional parameter space. \item We propose a heterogeneous MoE design framework and validate its effectiveness over homogeneous designs.
\item We apply HMMoE to audio-visual and video-text tasks, matching full fine-tuning performance with only 5-8\% of the parameters while outperforming existing PEFT methods.

\end{itemize}

\section{Related Work}
\subsection{Mixture of Experts}

Mixture of Experts (MoE) \cite{6797059,Eigen2013LearningFR,shazeer2017outrageously} is a neural network architecture that partitions layer parameters into discrete experts with distinct weights, activating only a subset of parameters during training and inference \cite{lewis2021base,li2022branchtrainmerge}. Related work \cite{Fedus2021SwitchTS,rajbhandari2022deepspeedmoe} has improved MoE performance by routing each input to a single expert, reducing computation while preserving model quality. This approach enhances computational efficiency compared to traditional methods where all network parameters are used. MoE has been widely applied in natural language processing \cite{hazimeh2021dselect,Komatsuzaki2022SparseUT,mustafa2022multimodal,zoph2022st,zhou2022mixture,lepikhin2020gshard} and computer vision \cite{chen2024eve,lin2024moe}, achieving significant success in visual-language tasks. Our method extends the performance of MoE by promoting modal interaction across various multimodal scenarios.

\subsection{Parameter-Efficient Fine-tuning}
Parameter-efficient fine-tuning \cite{houlsby2019parameterefficient} has become essential as model sizes increase. Strategies like Low-Rank Adaptation (LoRA) \cite{he2022unified,zhang2023lorafa} reduce parameters by adding trainable low-rank matrices, saving resources without extra inference cost. Adapters \cite{houlsby2019parameterefficient,pfeiffer2020adapterfusion} allow selective modification of pre-trained parameters, improving resource use without sacrificing performance. Prompt learning \cite{lester2021power} leverages task-specific prompts for fine-tuning with minimal parameters. Among these, low-rank adapters are particularly promising for resource savings and potential performance gains, inspiring our model to perform modal fusion in low dimensions to efficiently control fine-tuning parameters.

\section{The Method}

\subsection{Overview}

To demonstrate the functionality of the heterogeneous multi-modal mixture of experts adapter (HMMoE), we use a visual-audio task as an example. As shown in Figure \ref{fig:aaai}, the HMMoE module is inserted into the transformer structure after the feed-forward layer. For the $\ell^{th}$ layer, the module takes visual features $V^{\ell} \in \mathbb{R}^{B \times S_V \times D}$ and audio features $A^{\ell} \in \mathbb{R}^{B \times S_A \times D}$ as inputs and outputs fused features of the same dimension. Here, $B$ is the batch size, $S_A$ and $S_V$ are the sequence lengths for audio and visual inputs, and $D$ is the feature dimension.

The HMMoE structure has two expert groups: single-modal and multi-modal experts. The global router assigns weight factors to each group, while local routers determine the combination coefficients within each group. Multi-modal experts fuse visual and audio features, while single-modal experts retain and process each modality's unique information.

Regarding the overall structure, given the input video feature 
$V^{\ell}$ and audio feature $A^{\ell}$, the model's output can be expressed as follows:

\begin{equation}
V^{\ell+1}=\mathbf{G}_{m}^V\sum_{j=1}^M\mathbf{W}_{j}^V\cdot{E}_{j}\left(V^{\ell},A^{\ell} \right)+\mathbf{G}_{s}^V\sum_{j=1}^M\mathbf{W}_{j}^V\cdot{E}_{j}^{V}\left(V^{\ell}\right) 
\end{equation}
\begin{equation}
A^{\ell+1}=\mathbf{G}_{m}^A\sum_{j=1}^M\mathbf{W}_{j}^A\cdot{E}_{j}\left(A^{\ell},V^{\ell} \right)+\mathbf{G}_{s}^A\sum_{j=1}^M\mathbf{W}_{j}^A\cdot{E}_{j}^{A}\left(A^{\ell}\right) 
\end{equation}

Where $\mathbf{G}$ represents the weight given by the global router, $\mathbf{W}$ represents the weight given by the local router within the group, E represents a single single-modal or multi-modal expert, and M is the number of experts within each group.

\subsection{Heterogeneous Expert Group}
The heterogeneous expert groups are designed to enhance interaction between modalities at different perceptual dimensions. These groups are divided into multi-modal and single-modal experts. Multi-modal experts fuse features from both modalities using global attention and channel attention mechanisms, enabling cross-modal information transfer. Single-modal experts focus on extracting modality-specific information. This structure preserves the original modal information while facilitating multi-dimensional interaction between modalities.

\subsection{Multi-modal Expert}

The multi-modal expert facilitates the interaction and fusion of different modalities. To minimize parameter usage, we apply low-rank decomposition to map features to a smaller dimension, performing modal feature interactions in this reduced space. This method maintains model performance while significantly reducing parameters. We propose two multi-modal experts: one for cross-modal attention and another for channel-dimensional attention.

\subsubsection{Cross-modal Attention Expert}

The cross-modal attention expert facilitates modality interaction by capturing complex relationships. Given visual features \( V \in \mathbb{R}^{B \times S_V \times D} \) and audio features \( A \in \mathbb{R}^{B \times S_A \times D} \), both are projected to a lower-dimensional space \( r \) via \( \mathcal{W}_{down} \), resulting in \( \overline{V} \in \mathbb{R}^{B \times S_V \times r} \) and \( \overline{A} \in \mathbb{R}^{B \times S_A \times r} \).  For Audio-to-Visual attention, \( \overline{V} \) serves as a query (via \( \mathcal{W}_{q} \)), while \( \overline{A} \) provides key and value representations (via \( \mathcal{W}_{k} \) and \( \mathcal{W}_{v} \)). Attention weights, computed from the query-key dot product and softmax, weight the value to generate the output. The result, combined with the residual low-dimensional features, is up-projected back to the original dimension using \( \mathcal{W}_{up} \).

\begin{equation}
\overline{V} =\mathcal{F}_{relu}(V \cdot \mathcal{W}_{down})
\end{equation}
\begin{equation}
    V_{out}=\left( softmax\left( \frac{\overline{V}\mathcal{W}_{q} (\overline{A}\mathcal{W}_{k})^T}{\sqrt{d_V}}              \right)\overline{A}\mathcal{W}_{v} + \overline{A} \right)\cdot \mathcal{W}_{up}
\end{equation}

\subsubsection{Channel-Attention Experts}

In cross-modal tasks, global attention mechanisms may miss fine-grained modality-specific information. To address this, we introduce a channel-attention expert that focuses on the channel dimension, ensuring detailed modality information is preserved. The channel-attention expert processes two modal features, $V \in \mathbb{R}^{B\times S_V \times D}$ and $A \in \mathbb{R}^{B\times S_A \times D}$. For audio-to-video attention, the feature $A$ is averaged along the dimension $S_A$ and then multiplied element-wise by $V$. Then $V$ is projected into a lower-dimensional space using $\mathcal{W}_{down}\in \mathbb{R}^{D \times r}$ to minimize the parameters. The resulting feature is multiplied element-wise with the channel attention weights and residual connected to the original feature, producing the output $V{out}$. The entire process is described as follows:
 \begin{equation}
    Attn=Sigmoid\left(   AvgPool_{s}   \left( AvgPool_{s}(A)\cdot V \right) \right)
\end{equation}
\begin{equation}
    V_{out}=\mathcal{F}_{relu}(V \cdot \mathcal{W}_{down})\cdot \mathcal{W}_{up}\cdot(1+Attn)  
\end{equation}

\subsection{Single-modal Experts}

Over-relying on cross-modal information would hinder the model's ability to capture modality-specific features, degrading performance. To mitigate this, we introduce single-modal experts with simplified structures to preserve individual modality feature extraction. The single-modal expert follows the adapter design, consisting of two fully connected layers. For the input $V \in \mathbb{R}^{B\times S_V \times D}$, the feature dimension $D$ is reduced to a bottleneck dimension $r$ using a down-projection $\mathcal{W}{down} \in \mathbb{R}^{D \times r}$, and then restored to the original dimension with an up-projection $\mathcal{W}{up} \in \mathbb{R}^{r \times D}$. The process is represented as:
\begin{equation}
    V_{out} = V + \mathcal{F}_{relu}(V \cdot \mathcal{W}_{down})\cdot \mathcal{W}_{up}
\end{equation}
\subsection{Routing Method}

The routing method aims to select the most suitable experts for processing input features. Our model employs two types of routers: the global router assigns weights to expert groups, while the local router selects the top-k experts within each group.

\subsubsection{Global Router}  
The global router assigns weight coefficients to different expert groups to leverage their strengths. Instead of fixed weights, we use a learnable weight-allocation mechanism. The global routing weight, \(G_{s,m}(x)\), is computed as:
\begin{equation}
G_{s,m}(x) = \text{SoftMax}(W_{gr}(x))    
\end{equation}

where \(W_{gr}\) is a set of learnable linear mappings. This approach allows the model to dynamically adjust its preferences for different expert levels.

\subsubsection{Local Router}  
Local routers are responsible for selecting the most appropriate experts within each group. For a multimodal expert group \(\overline{E}_m = [E_1, E_2, \dots, E_N]\), where \(N\) represents the total number of experts, the weight for each expert is computed as \(P(x)\). The top-k experts, based on the highest probability, process each feature, and the weighted sum is calculated as:
\begin{equation}
P(x)_i = \frac{e^{W^{lr}_i x}}{\sum_{j=1}^{N} e^{W^{lr}_j x}} 
\label{eq9}
\end{equation}
\begin{equation}
Group(x) = \text{TopK}(P(x)_i \cdot E(x)_i)    
\end{equation}

Through the joint operation of the global and local routers, the model ensures the efficient and optimal allocation of experts for processing the input features.



\section{Experiments}

\begin{table*}[t]
    \centering
        \caption{Performance comparison of visual-audio tasks based on Swin-T and HT-SAT encoders: a comparison with traditional methods with equal parameters.}
    \resizebox{\textwidth}{!}{  
    \begin{tabular}{lcccccccccc}
        \toprule
        &  &\textbf{AVE}&\multicolumn{2}{c}{\textbf{AVVP}}&\multicolumn{4}{c}{\textbf{AVQA}}&\multicolumn{2}{c}{\textbf{AVS-S4}}   \\
        
        \textbf{Method} & \textbf{Parameters (M) } & \textbf{Acc} & \textbf{seg-level} & \textbf{ event-level} & \textbf{AQ} & \textbf{VQ} & \textbf{AVQ} & \textbf{Avg} & \textbf{mIoU} & \textbf{F} \\
        \midrule
        \textit{Full-finetune} & \textit{313(100\%)} & 82.2 & 52.8 & 46.1 & 77.4 & 81.9 & 70.7 & 74.8 & 80.9 & 89.2 \\
        
        Lora &20(6.3\%) & 79.8 & 52.6 & 45.9 & 75.4 & 81.3 & 70.5 & 74.3 & 79.8 & 88.1 \\
        Lora-FA &20(6.3\%) & 79.5 & 52.5 & 46.0 & 75.1 & 80.9 & 70.7 & 74.7 & 79.2 & 87.9 \\
        Series-Adapter &20(6.3\%) & 79.9 & 52.0 & 45.9 & 76.3 & 81.9 & 70.2 & 74.2 & 80.2 & 88.6 \\
        Parallel-Adapter &20(6.3\%) & 80.2 & 52.3 & 45.3 & \textbf{76.9} & 81.7 & 71.1 & 74.9 & 80.1 & 88.8 \\
        Ours &20(6.3\%) & \textbf{81.1} & \textbf{53.4} & \textbf{46.8} & 76.7 & \textbf{82.4} & \textbf{71.3} & \textbf{75.1} & \textbf{80.9} & \textbf{89.3} \\
        \bottomrule
    \end{tabular}
    }

    \label{tab:audio}
\end{table*}

\subsection{Experiment Setup}

In this section, we describe our model training procedure. We integrate the HMMoE module into the encoder layer of a pre-trained multi-modal model, initializing both single-modal and multi-modal experts. During training, we freeze the original model parameters and train only the HMMoE layers and the classification head. For comparison, we also evaluate traditional PEFT methods, including series-adapter \cite{houlsby2019parameterefficient}, parallel-adapter \cite{he2021towards}, LoRA \cite{hu2021lora}, and LoRA-FA \cite{zhang2023lorafa}.

We evaluate our method on both visual-audio and text-visual tasks. For visual-audio tasks, we use the pre-trained Swin-T \cite{liu2021swin} and HT-SAT \cite{chen2022hts} models as encoders, performing experiments on AVE \cite{tian2018audiovisual}, AVVP \cite{tian2020unified}, AVQA \cite{li2022learning}, and AVS \cite{zhouavsbench} tasks. For text-visual tasks, we implement MSVD and MSRVTT \cite{xu2016msr} datasets using the pre-trained VALOR \cite{chen2023valor} model, a dual-tower encoder that processes multi-modal information. Additionally, we test our method on the VQA and NLVR tasks using the VLMO \cite{bao2022vlmo} model, which is based on the MOE architecture. All experiments were conducted on A800 GPUs, using the same training settings as the base model. Further experimental details can be found in the supplementary materials.




\subsection{Implementation Details}

\subsubsection{Visual-Audio Tasks}
We integrate the HMMoE module into the Swin-T and HT-SAT models for various tasks. For AVE, our module works with CMBS, and accuracy is used as the metric. For AVVP, it is combined with MGN and evaluated using segment-level and event-level metrics across audio, visual, and audio-visual events. For AVS, the module integrates with the AVS model, assessed by mIoU and F-score. In AVQA, it is incorporated into the ST-AVQA framework, using answer accuracy as the metric.

\subsubsection{Text-Visual Tasks}
We use the VALOR model for video QA tasks on MSRVTT-QA and MSVD-QA datasets, evaluated by QA accuracy. For vision-language classification, we leverage the VLMo model on NLVR2 and VQA2 datasets, with QA accuracy as the metric.
\begin{table}[h]
    \centering
    
    \caption{     Performance of text-visual tasks based on VALOR: comparison with traditional methods with equal parameters.}
      \resizebox{\linewidth}{!}{
    \begin{tabular}{l@{\hskip 10pt}c@{\hskip 5pt}c@{\hskip 5pt}c@{\hskip 5pt}}
        \toprule
        \textbf{Method} & \textbf{Parameters(M)} & \textbf{MSRVTT} & \textbf{MSVD} \\
            \midrule
       \textit{Full-finetune} & \textit{315(100\%)} & 44.5 & 54.9 \\
        Lora & 16(5.1\%) & 43.7 & 53.5 \\
        Lora-FA & 16(5.1\%) & 43.0 & 53.1 \\
        Series-Adapter & 16(5.1\%) & 43.6 & 54.1 \\
        Parallel-Adapter & 16(5.1\%) & 44.1 & 54.2 \\
        \textbf{Ours} & 16(5.1\%) & \textbf{45.2} & \textbf{55.6} \\
        \bottomrule
    \end{tabular}
    }\label{tab:VALOR}
\end{table}
\begin{table}[h]
  \centering
  \caption{Performance on text-visual tasks based on VLMO: comparison with traditional methods with equal parameters.}
    \resizebox{\linewidth}{!}{
   \begin{tabular}{lccc}    \toprule
    \textbf{Model} & \textbf{Parameters(M)} & \textbf{VQA} & \textbf{NLVR} \\    \midrule

    \textit{Full-finetune} & \textit{360(100\%)} & 76.2 & 82.7 \\
    Lora & 19(5.3\%) & 73.8 & 80.9 \\
    Lora-FA & 19(5.3\%) & 73.6 & 80.5 \\
    Series-Adapter & 19(5.3\%) & 74.4 & 81.1 \\
    Parallel-Adapter & 19(5.3\%) & 74.6 & 81.4 \\
    \textbf{Ours} & 19(5.3\%) & \textbf{75.2} & \textbf{82.2} \\
    \bottomrule
  \end{tabular}
  }
  \label{tab:vlmo}
\end{table}

\subsubsection{Comparison setting}
In these tasks, our HMMoE module was configured with single-modal, cross-modal,el-attention expert groups, each containing two experts, with the rank (r) of the experts set to 32 and the toand the router method. In order to make a fair comparison with traditional fine-tuning methods including Lora, Lora-FA, serial adapter, and parallel adapter, we adjust the value of the low-rank mapping dimension r to ensure that the number of parameters used by various methods is consistent.

\subsection{Main Results}
\subsubsection{Performance Comparison}

Our HMMoE method outperforms existing approaches in visual-audio tasks, as shown in Table \ref{tab:audio}. It achieves the highest accuracy and overall performance across all tasks. In the AVE task, our model achieves the best accuracy, and in the AVVP and AVS-S4 tasks, it leads in both accuracy and efficiency. This success is due to our model’s effective strategy of combining Swin-T and HT-SAT encoders, which enhances its ability to capture the relationship between audio and visual information.

In text-visual tasks, our HMMoE module also outperforms other methods. On the VALOR benchmark, it improves performance on MSRVTT and MSVD by 0.7 percentage points. On the VLMO benchmark, our model surpasses the closest competitor by 0.6 percentage points on VQA and 0.8 percentage points on NLVR. This improvement is driven by our model's ability to effectively integrate text and visual information, offering superior cross-modal feature fusion and better generalization compared to traditional PEFT methods.

\begin{figure}[ht]
    \centering
    \begin{subfigure}[b]{0.49\linewidth}  
        \centering
        \includegraphics[width=\textwidth]{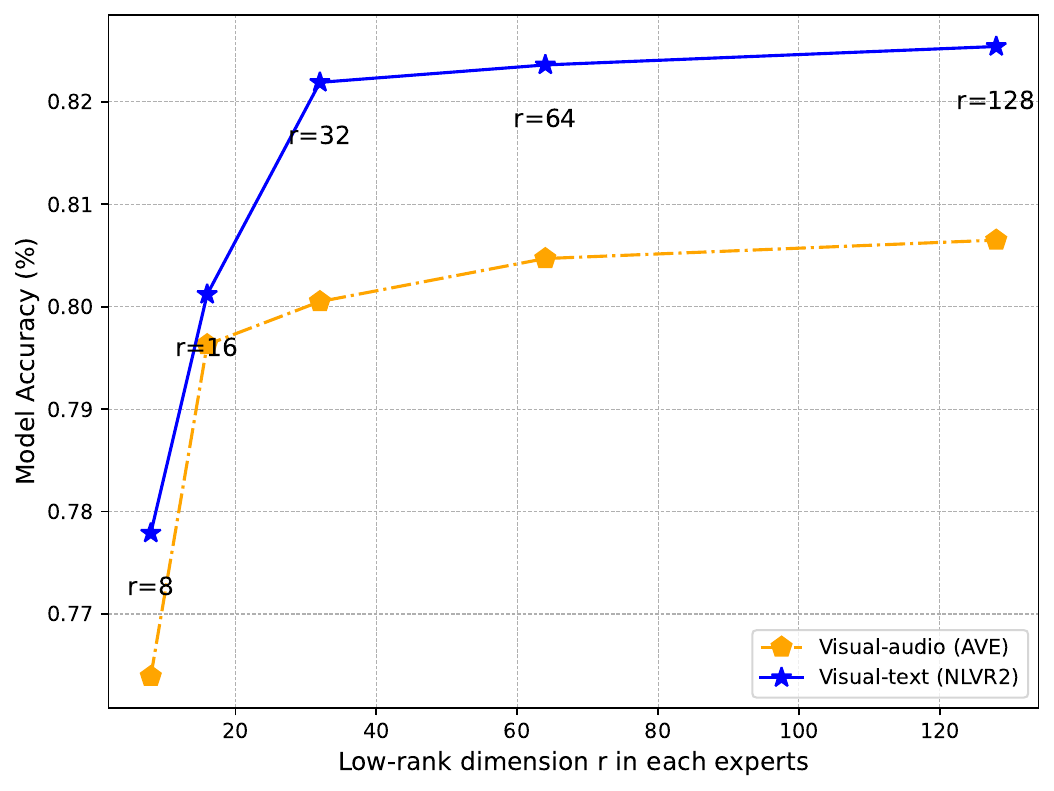}
        \caption{Rank ablation}
        \label{fig:example}
    \end{subfigure}
    \hfill  
    \begin{subfigure}[b]{0.49\linewidth}  
        \centering
        \includegraphics[width=\textwidth]{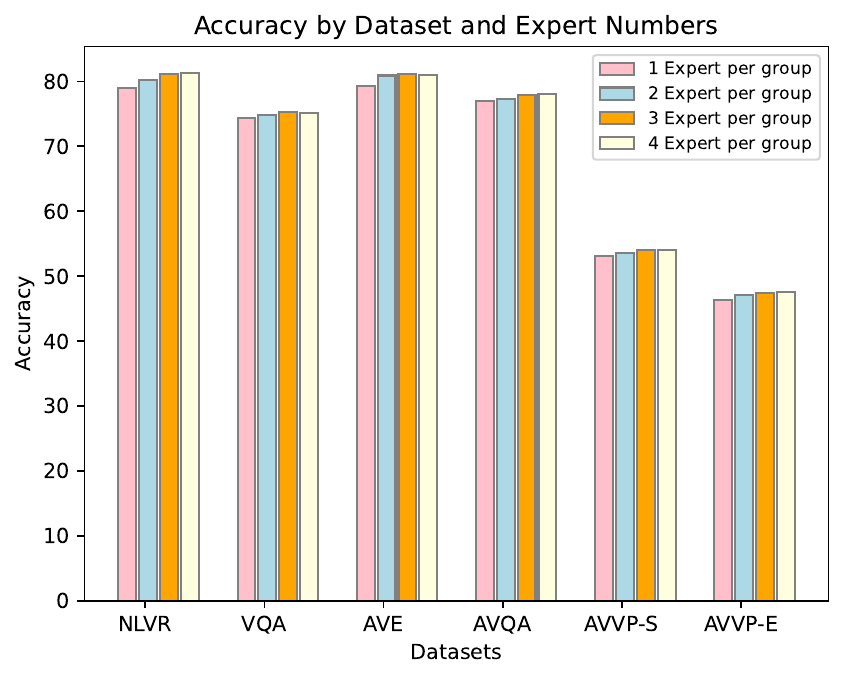}
        \caption{Expert number ablation}
        \label{fig:number}
    \end{subfigure}
    \caption[width=0.4\textwidth]{The graph (a) shows the relationship between the low-rank dimension r and model performance, and the graph (b) shows the relationship between the number of experts in each group and model performance.}
    \label{fig:combined}
\end{figure}

\begin{table}[t]
  \centering
  \small 
  \setlength{\tabcolsep}{4pt} 
    \caption{ Ablation results of the expert module in text-visual tasks (Single for single-mode experts, Cross for cross-attention experts, and Channel for channel attention experts).}
    \begin{tabular}{cccccccc}
    \toprule
     \multicolumn{3}{c}{\textbf{Module}} &\textbf{NLVR} & \textbf{VQA} & \textbf{MSVD} & \textbf{MSRVTT}  \\
    \textbf{Single}  & \textbf{Cross}  & \textbf{Channel} & Acc & Acc & Acc & Acc\\ \midrule
    \checkmark      & - & - & 80.4 & 74.4& 54.1&43.6\\
    \checkmark       & \checkmark & -& 81.0 & 75.1&54.2 &44.6\\
    \checkmark      & \checkmark & \checkmark & \textbf{82.2} & \textbf{75.2}&\textbf{55.6}&\textbf{45.2} \\
    \bottomrule
    \end{tabular}%

  \label{tab:abalation study}%
\end{table}%


\subsection{Ablation Study}
\subsubsection{Expert module ablation} 
To demonstrate the impact of single-modal and multi-modal experts on performance, we conducted an ablation study on the type of expert used in downstream tasks (NLVR2, VQA, MSVD-QA, MSRVTT-QA). As shown in Table \ref{tab:abalation study}, incorporating cross-attention experts improves performance over using only single-modal experts. Further gains in fine-tuning performance were observed with the addition of channel-attention experts, which enhance targeted dimension extraction. These results highlight the robustness of expert combinations in downstream tasks.

\begin{figure}[b]
    \centering
    \begin{subfigure}[b]{0.48\linewidth}
        \centering
        \includegraphics[width=\linewidth]{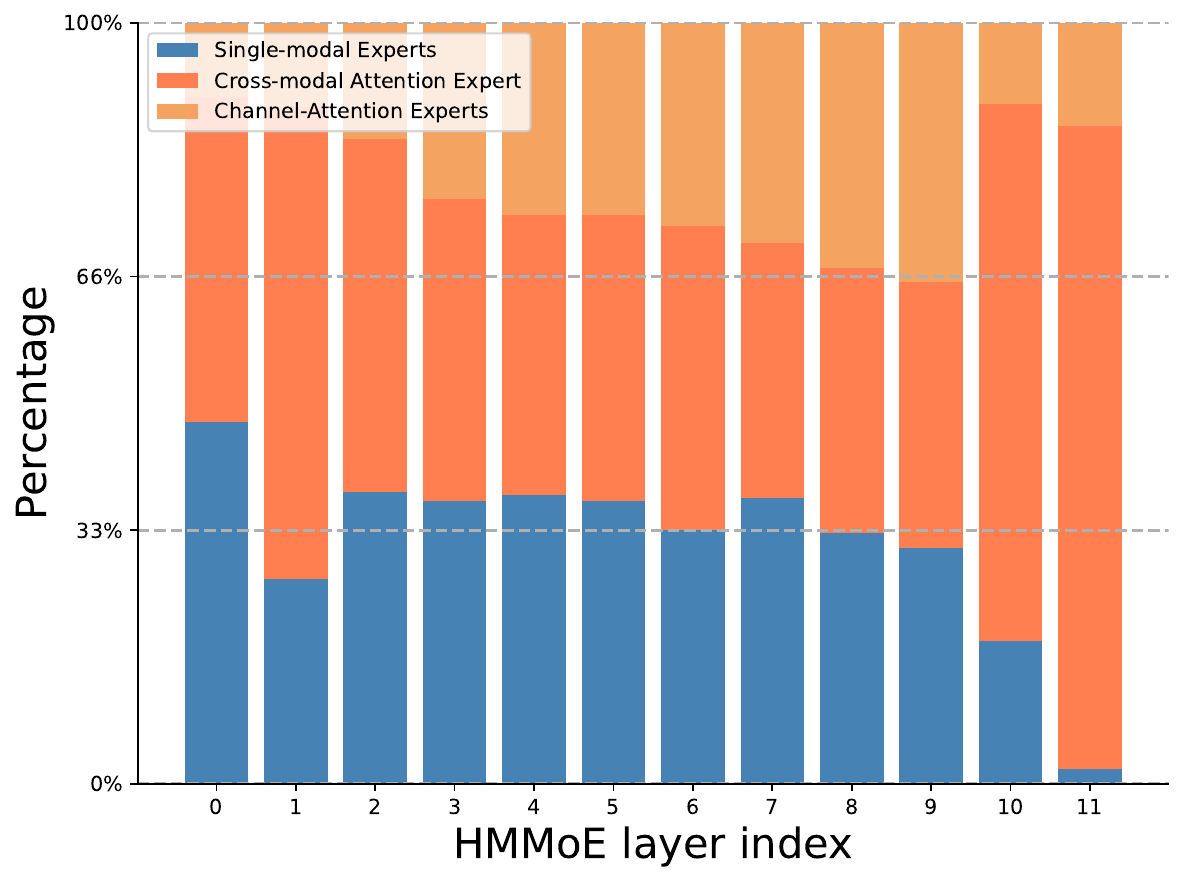}
        \caption{Visual-audio task}
        \label{fig:heat1}
    \end{subfigure}
    \hfill
    \begin{subfigure}[b]{0.48\linewidth}
        \centering
        \includegraphics[width=\linewidth]{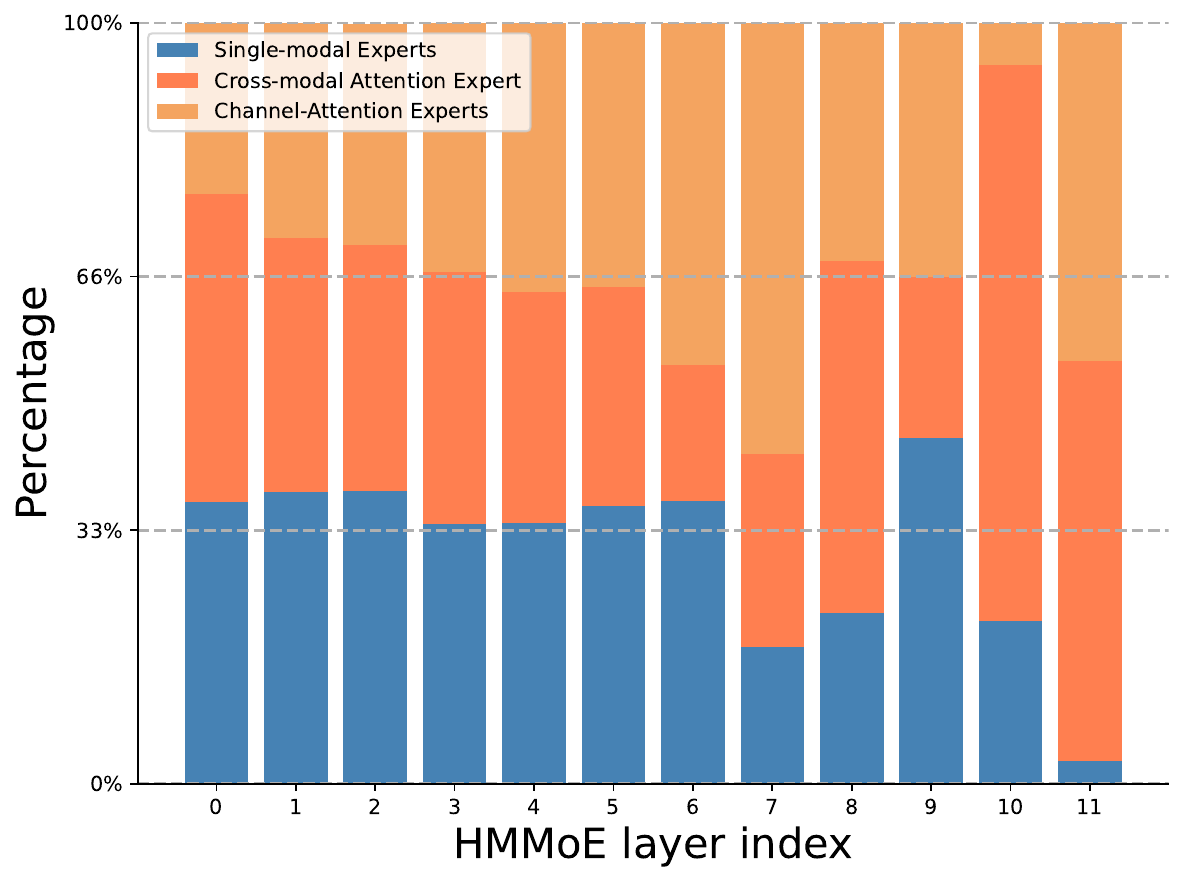}
        \caption{Text-visual task}
        \label{fig:heat2}
    \end{subfigure}
    \caption{Distribution of expert utilization across different layers for each expert type.}
    \label{fig:heat_comparison}
\end{figure}


\subsubsection{Low-dimention rank ablation}
In our HMMoE method, the dimension r of the low-rank mapping plays a crucial role in reducing the original feature dimension to a lower-dimensional space. Decreasing r can help reduce the model's training parameters, but setting r too low may lead to a significant loss of original feature information, resulting in degraded performance. As shown in Figure \ref{fig:example}, the overall performance of the model increases rapidly as r increases. However, once r reaches around 32, further increases in r yield diminishing returns in performance improvement. Therefore, it is important to select r within a reasonable range: setting r too low can result in the loss of critical feature details while setting it too high can lead to unnecessary resource consumption without significant performance gains.

\subsubsection{Experts number ablation}
In general, the performance of the model increases as the number of experts in each group increases. As shown in Figure \ref{fig:number}, the rate of performance improvement diminishes with the addition of more experts. Since our goal is to fine-tune the entire multimodal model with minimal overhead, it is preferable to limit the number of experts to a low level, such as 2 or 3, to maintain performance without significantly increasing the model’s complexity. This approach allows us to effectively balance performance gains with the number of parameters, ensuring an efficient trade-off.



\subsection{Heterogeneous Effect Analysis} 
To validate the effectiveness of our heterogeneous expert module, we compared it with models using multiple single-expert combinations. The results show that models with mixed heterogeneous experts outperform the others, not only due to the increased number of experts but also because of the innovative heterogeneous structure. The integration of single-modal information into other modalities via cross-attention and channel-attention mechanisms significantly improves modality fusion, while maintaining a low parameter count.


\begin{table}[h]
\centering
 \caption{
    Evaluation of heterogeneous experts' efficiency across NLVR, VQA and AVE tasks.}
\begin{tabular}{ccccc}
\toprule
& \multicolumn{1}{c}{\textbf{Single Expert}} & \multicolumn{1}{c}{\textbf{Cross Expert}} & \multicolumn{1}{c}{\textbf{Channel Expert}} & \textbf{Acc} \\
\midrule
\multirow{2}{*}{NLVR} & 6 & - & & 81.1 \\
& 2 & 2 & 2 & \textbf{82.2} \\
\midrule
\multirow{2}{*}{VQA} & 4 & - & & 74.9 \\
& 2 & 1 & 1 & \textbf{75.2} \\
\midrule
\multirow{2}{*}{AVE} & 3 & - & & 79.8 \\
& 1 & 1 & 1 & \textbf{81.0} \\
\bottomrule
\end{tabular}
  \label{tab:pretrained_models}
\end{table}

In the HMMoE module, expert selection varies across transformer layers. Lower-level layers maintain a balanced expert selection, while higher-level layers prioritize multi-modal experts. As shown in Figure \ref{fig:heat_comparison}, the model favors channel-attention and cross-attention experts, indicating that incorporating cross-modal information enhances classification performance.



\section{Conclusion}
In conclusion, we introduce a novel Heterogeneous Multi-modal Mixture of Experts Adapter (HMMoE) to address the limitations of existing parameter-efficient fine-tuning methods in multi-modal  models. Our approach extends the input of each expert from a single modality to multiple modalities, enabling effective cross-modal interactions within each expert. By mapping inputs to a low-rank space for interaction and subsequently back to their original dimensions, our method facilitates efficient gradient adjustments of the frozen pre-trained model parameters based on collaborative multi-modal features. 
Additionally, we have transitioned from the traditional single-expert structure to a heterogeneous expert framework that integrates various interaction types, including cross-attention experts and channel-attention experts. This more diverse architecture allows our model to better capture and process the intricate relationships within multi-modal data. The experimental results highlight the effectiveness and advantages of our proposed module, showing significant improvements in managing complex multi-modal scenarios.

\bibliographystyle{IEEEbib}
\bibliography{icme2025_template_anonymized}

\end{document}